\RequirePackage{amsmath}
\documentclass[runningheads,a4paper]{llncs}

\usepackage{amssymb}
\usepackage{graphicx}

\usepackage{url}
\urldef{\mailsa}\path|{tietz, alpay, twiefel, wermter}@informatik.uni-hamburg.de|
\newcommand{\keywords}[1]{\par\addvspace\baselineskip
\noindent\keywordname\enspace\ignorespaces#1}

\usepackage{tabularx}
\usepackage{color}
\usepackage{hyperref}
\usepackage[caption=false]{subfig}
\usepackage{units}

\definecolor{enc}{RGB}{90,160,44}
\definecolor{dec}{RGB}{255,204,0}

\definecolor{dark-blue}{rgb}{0.15,0.15,0.4}
\definecolor{medium-blue}{rgb}{0,0,0.5}\definecolor{dark-blue}{rgb}{0.15,0.15,0.4}
\definecolor{medium-blue}{rgb}{0,0,0.5}
\hypersetup{
	colorlinks, linkcolor={black}, citecolor={dark-blue}, urlcolor={medium-blue}
}

\newcommand{\myheader}{In: Proc. of the 26th Int. Conference on Artificial Neural Networks (ICANN), Alghero, Italy, September 11-14, 2017\\
The final publication is available at Springer via  \url{http://dx.doi.org/10.1007/978-3-319-68600-4_1}\\}
\usepackage{atbegshi,picture}
\usepackage{datetime}                    
\usepackage{lastpage}                    

\newcommand{\myleftstd}{1.1in}
\ifthenelse{\isodd{\thepage}}
   {\newcommand{\myleftmargin}{\oddsidemargin+\myleftstd}}
   {\newcommand{\myleftmargin}{\evensidemargin+\myleftstd}}

\AtBeginShipoutNext {
		\AtBeginShipoutUpperLeft{%
  \put(\dimexpr 0\myleftmargin\relax,-1.2cm){\parbox{1.3\textwidth \myleftmargin }{\footnotesize\sffamily\noindent{\centering \myheader \hfill}}}%
}}

\begin{document}

\mainmatter  

\title{Semi-Supervised Phoneme Recognition with Recurrent Ladder Networks}

\titlerunning{Semi-Supervised Phoneme Recognition with Recurrent Ladder Networks}

%
%
\author{Marian Tietz\and
Tayfun Alpay\and
Johannes Twiefel\and
Stefan Wermter
}
\authorrunning{Semi-Supervised Phoneme Recognition with Recurrent Ladder Networks}

\institute{Knowledge Technology Institute,
Department of Informatics,
Universit\"at Hamburg,
Vogt-K\"olln-Str.\ 30,
22527 Hamburg, Germany
\mailsa\\
\url{http://www.informatik.uni-hamburg.de/WTM/}}

%
%

\toctitle{Semi-Supervised Phoneme Recognition with Recurrent Ladder Networks}
\tocauthor{Marian Tietz, Tayfun Alpay, Johannes Twiefel, Stefan Wermter}
\maketitle

\begin{abstract}
Ladder networks are a notable new concept in the field of semi-supervised learning by showing state-of-the-art results in image recognition tasks while being compatible with many existing neural architectures. We present the recurrent ladder network, a novel modification of the ladder network, for semi-supervised learning of recurrent neural networks which we evaluate with a phoneme recognition task on the TIMIT corpus.
Our results show that the model is able to consistently outperform the baseline and achieve fully-supervised baseline performance with only 75\% of all labels which demonstrates that the model is capable of using unsupervised data as an effective regulariser.
\end{abstract}

\section{Introduction}
\label{sec:Introduction}

There is no doubt that the recent success of deep learning is tied to the rising availability of labelled data.
While tasks such as image or text classification have greatly benefited from this availability, there are still a number of domains, e.g. speech recognition, where the majority of the research community has no free access to large amounts of labelled data.
One promising approach towards this problem is semi-supervised learning where models trained with \textit{labelled} data can be further improved by training with \textit{unlabelled} data.

Recent methods, such as graph-supported training~\cite{liu2013graph}, sparse autoencoders (\cite{dhaka2016semi};~SSSAE) and especially the Ladder Network (LN)~\cite{rasmus2015semi}, a stacked Denoising Autoencoder (DAE) with shortcut connections, show promising results for semi-supervised training of feed-forward neural networks.
The LN has been shown to deliver state-of-the-art results in semi-supervised image classification while still being compatible with many existing feed-forward neural networks~\cite{rasmus2015semi}.

However, this novel architecture has not yet been explored on more complex sequential tasks, such as speech recognition, where Recurrent Neural Network (RNN) architectures, like Gated Recurrent Units (GRU;~\cite{cho2014learning}), are the current state of the art. We therefore propose a novel Recurrent Ladder Network (RLN) architecture and evaluate it on the TIMIT phoneme recognition benchmark~\cite{garofolo1993darpa}.
We introduce a novel recurrent layer for the LN decoder in order to find better-suited abstractions for semi-supervised learning and test two noise injection schemes tailored to support recurrent dynamics to increase the regularising nature of the RLN.
Our results show that after hyper-parameter optimization the model is able to significantly outperform the baseline in all experiments using unsupervised data as a regulariser and achieves fully-supervised baseline performance while training only on 75\% of the labelled data.

\section{The Ladder Network Architecture}
\begin{figure}
    \centering
    \hspace{-2em}
	\includegraphics[width=.55\textwidth]{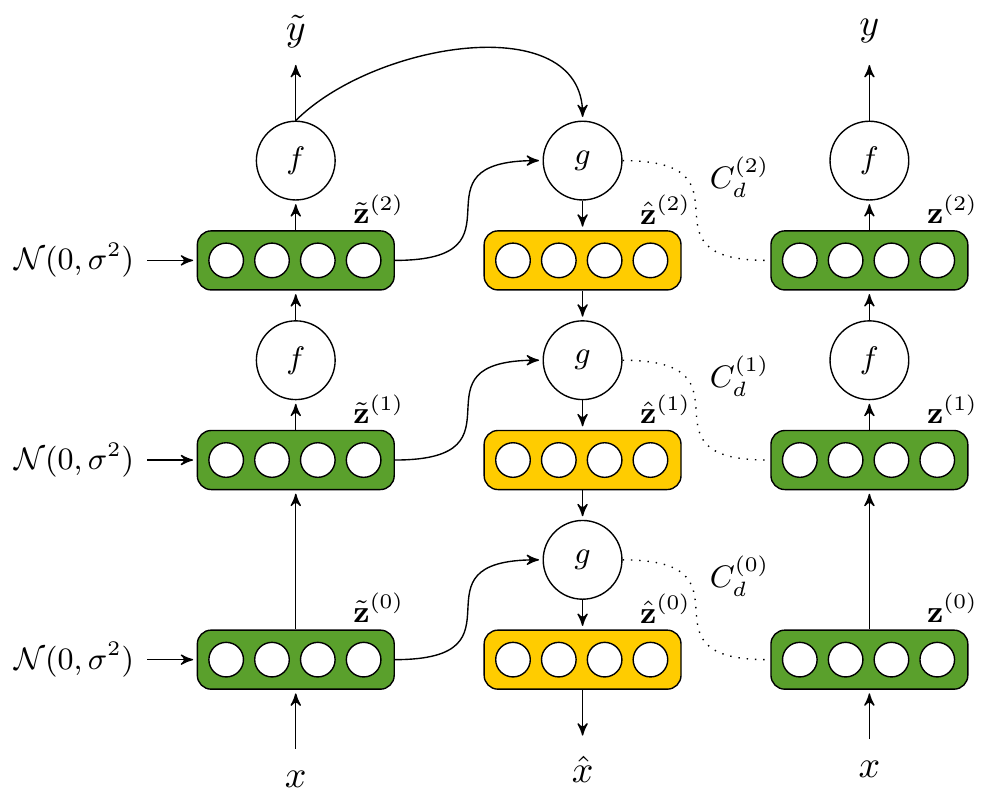}
    \caption{Illustration of the non-recurrent LN architecture with one hidden and one output layer. The encoder and decoder paths are highlighted in \textcolor{enc}{\textbf{green}} and \textcolor{dec}{\textbf{yellow}}, respectively.}
    \label{fig:ln}
\end{figure}

The basic idea of the LN architecture~\cite{rasmus2015semi}, depicted in Fig. \ref{fig:ln}, is to make autoencoders more expressive by adding shortcut connections from the encoder to the decoder. Each decoder layer is then able to combine the preactivation of the encoder layer with the reconstruction of the previous decoder layer by means of a combinator function $g(\cdot,\cdot)$.
Therefore, the encoder does not have to carry all reconstruction information since the shortcuts can compensate for it.
Since the shortcuts allow perfect reconstruction by simply copying the encoder input to the decoder output, Gaussian noise $\mathcal{N}(0,\sigma^2)$ is added to prevent the direct usage of these short-circuits and enforce learning in the intermediate layers, i.e. we use a denoising autoencoder. To ensure that the noise can be removed, the decoder's (noisy) reconstruction $\hat{\textbf{z}}^{(l)}$ is compared to the encoder's (clean) preactivation $\textbf{z}^{(l)}$ and added to the unsupervised objective function:
\begin{equation}
    C_{\text{DAE}} = \sum_{l}^{n} \lambda_{l} C^{(l)}_{d} ~~\text{with}~~
    C^{(l)}_d = \|~ \textbf{z}^{(l)} - \hat{\textbf{z}}^{(l)} \|^{2} ~,
\end{equation}
where $n$ is the total amount of layers, $\textbf{z}^{(l)}$ is the preactivation vector of the $l$-th encoder layer without noise and $\hat{\textbf{z}}^{(l)}$ the $l$-th decoder layer reconstruction from noisy input.
The hyper-parameter $\lambda_i$ controls the targeted similarity between the encoder and decoder layers and prevents short-circuits by punishing direct copies of the noisy data by weighting the difference between the layers.
For semi-supervised learning the encoder path is also used for the supervised task, i.e. its output is evaluated with a supervised objective function $C_{\text{sup}}$ and combined with the unsupervised objective function $C_{\text{DAE}}$: $C_{\text{semsup}} = C_{\text{sup}} + C_{\text{DAE}}$.
When using the encoder in a supervised task the shortcuts help with reconstruction as the needed information may also be retrieved over the shortcuts~\cite{rasmus2015semi}.

The combinator function $g(\cdot,\cdot)$ models $p(\textbf{z}^{(l)}\mid \textbf{z}^{(l+1)})$ and is responsible for creating the reconstruction of the $l$-th layer $\hat{\textbf{z}}^{(l)}$ with the help of the reconstruction of the previous layer $\hat{\textbf{z}}^{(l+1)}$ and the shortcut value of the $l$-th layer $\tilde{\textbf{z}}^{(l)}$, i.e., $\hat{\textbf{z}}^{(l)}  = g(\tilde{\textbf{z}}^{(l)},\hat{\textbf{z}}^{(l+1)})$.
The function may attempt to remove the noise from $\tilde{\textbf{z}}^{(l)}$ with the help of the previous reconstruction, infer the inverse mapping $\hat{\textbf{z}}^{(l+1)} \rightarrow \hat{\textbf{z}}^{(l)}$ or do a combination of both.

\section{Recurrent Ladder Networks}
\label{sec:rln}
In this section, we will elaborate our modelling choices for the RLN. In order to extend the original LN to support recurrence in the encoder, both the noise injection scheme and the decoder have to be adapted since recurrent layers use additional context layers.
Overall, we are proposing two noise injection methods and two decoder variants (see Fig. \ref{fig:recg}). Our supervised baseline model will be the encoder of the RLN since it encodes the task closely to the full RLN but has no means of using unsupervised data. The resulting six model combinations are No-Decoder with Feed-Forward Noise (ND-FFN), No-Decoder with Recurrent Noise (ND-RN), Recurrent Decoder with Feed-Forward Noise (RD-FFN) and Recurrent Noise (RD-RN) as well as a Feed-Forward Decoder with Feed-Forward Noise (FFD-FFN) and Recurrent Noise (FFD-RN).

\subsection{Noise Injection}
\label{sec:rln_noise}

In the \textit{feed-forward} case, noise is applied directly to the preactivations so that the output of the layer and the shortcut are affected, i.e. $\tilde{\textbf{z}} = W\tilde{\textbf{x}} + \textbf{n}$ with $\textbf{n} \sim \mathcal{N}(0,\sigma^2)$.
This would, however, introduce noise into the context memory of recurrent layers even \textit{after} receiving the noisy output from the previous layer, effectively amplifying the noise even further. Therefore, we apply noise only to the preactivation and the shortcut without direct perturbation of the context memory. A hidden layer $\textbf{h}_t$ and its noisy counterpart $\tilde{\textbf{h}}_t$ are therefore updated as follows:
\begin{align}
    \textbf{h}_{t} &= f(\textbf{z}_{t}) = f(W\tilde{\textbf{x}}_{t} + U\textbf{h}_{t-1}),\\
    \tilde{\textbf{h}}_{t} &= f(\tilde{\textbf{z}}_{t}) = f(\textbf{z}_{t} + \textbf{n}),
\end{align}
where $f(\cdot)$ is the activation function, $\textbf{x}_t$ the input, $W$ the input weight matrix, and $U$ the hidden-to-hidden weights, updated at each time step $t$.

This noise injection method will be referred to as \textit{recurrent noise} from here on. Another method of noise injection that we tested, referred to as \textit{feed-forward noise}, is to not inject additional noise at the recurrent layer, i.e. feed-forward layers will be injected with noise but recurrent layers will not.

\subsection{Recurrent Decoder}
\label{sec:rln_decoder}
\begin{figure}
    \centering
	\def\myfigheight{14em}
    \subfloat[][FFN]{
		\includegraphics[height=\myfigheight]{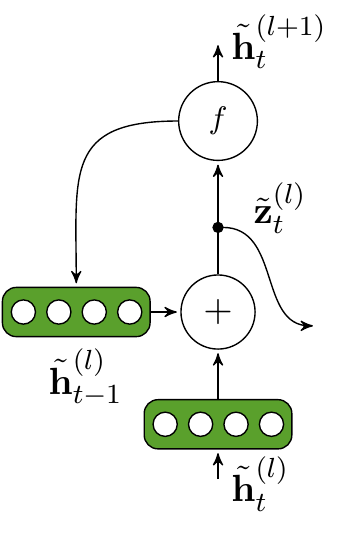}
		\label{fig:ffn}
	}
	\hfill
    \subfloat[][RN]{
		\includegraphics[height=\myfigheight]{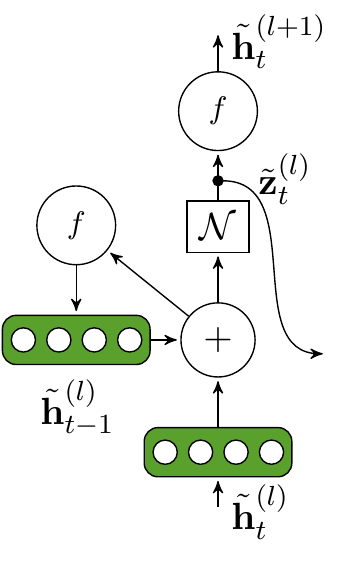}
		\label{fig:rn}
    }
	\hfill
    \subfloat[][RD]{
        \centering
		\includegraphics[height=\myfigheight]{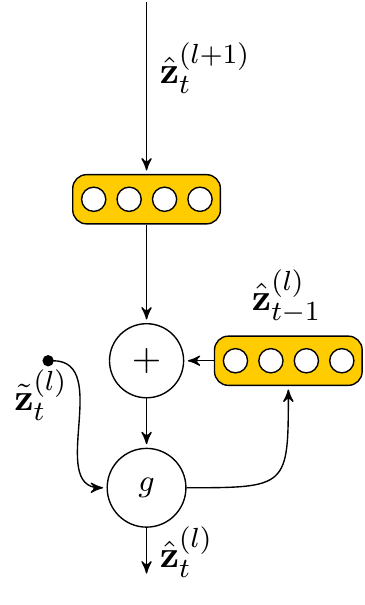}
		\hspace{-2em}
		\label{fig:rd}
	}
	\hfill
    \subfloat[][FFD]{
        \centering
		\includegraphics[height=\myfigheight]{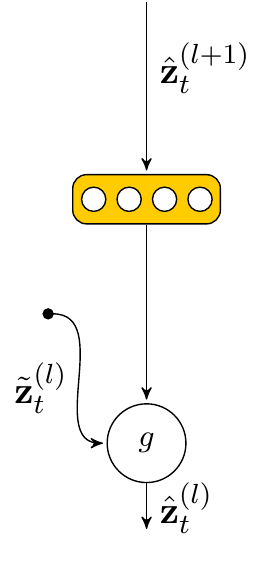}
		\label{fig:ffd}
    }%
	\hfill
    \caption{
    Overview of (a) feed-forward noise (FFN) and (b) recurrent noise (RN) injection schemes for the encoders (\textcolor{enc}{\textbf{green}}) introduced in subsection \ref{sec:rln_noise} as well as (c) recurrent decoder (RD) and (d) feed-forward decoder (FFD) layouts (\textcolor{dec}{\textbf{yellow}}) introduced in subsection \ref{sec:rln_decoder}. Combining all encoder and decoder layouts gives a total of six model variants including the two no-decoder (ND) baselines ND-RN and ND-FFN.}
    \label{fig:recg}
\end{figure}

The decoder path in an autoencoder models the inverse information flow of the encoder path. We propose two modelling options for the decoder path in an RLN. The first (Fig. \ref{fig:rd}) is a recurrent layer with $g(\cdot,\cdot)$ as activation function:
\begin{align}
    \textbf{u}^{(l)}_{t} &= V\hat{\textbf{z}}^{(l+1)}_{t} + O\hat{\textbf{z}}^{(l)}_{t-1},\\
    \hat{\textbf{z}}^{(l)}_{t} &= g(\tilde{\textbf{z}}^{(l)}_{t}, \textbf{u}^{(l)}_{t}),\label{eq:foo}
\end{align}
where $V$ are the input weights, $O$ the hidden-to-hidden weights, $\textbf{u}^{(l)}_t$ the preactivation of the recurrent decoder and $\tilde{\textbf{z}}^{(l)}_{t}$ the noisy preactivation of the $l$-th encoder layer at time-step $t$ from the shortcut.
The second modelling option is to simply use a feed-forward network (Fig. \ref{fig:ffd}) in the decoder~\cite{rasmus2015semi}.

Batch normalisation is heavily used in the LN both for normalisation of the layer-wise reconstruction cost and for normalisation of layer activations. It was considered problematic with recurrent networks until the introduction of recurrent batch normalisation~\cite{cooijmans2016recurrent}. Since it potentially requires tuning of another hyper-parameter we decided to model the RLN without batch normalisation with the exception of the layer-wise reconstruction cost function $C^{(l)}_{d}$ which is computed exactly as described by Rasmus et al.~\cite{rasmus2015semi}.

\section{Experiments}

We evaluate the RLN on the TIMIT phoneme recognition benchmark~\cite{garofolo1993darpa}, a widely used test corpus which allows comparing our architecture to previous approaches.
The audio samples of the corpus are reduced in dimensionality by using libROSA\footnote{\url{https://librosa.github.io}} to compute 13 Mel Frequency Cepstral Components (MFCC)~\cite{davis1980comparison} and their first and second derivative with 20\unit{ms} frames and 10\unit{ms} frame skip, similar to related work~\cite{dhaka2016semi}. The 39-dimensional feature vectors are normalised to have zero mean and unit variance.
We grouped easily confused phonemes of the English phoneme alphabet as described by Halberstadt~\cite{halberstadt1998heterogeneous} resulting in 39 phoneme classes to predict.

We use Connectionist Temporal Classification (CTC)~\cite{graves2006connectionist} for the supervised cost $C_{\text{sup}}$ to solve the problem of label alignment.
Phoneme Error Rate (PER) is used for evaluation and computed using the Levenshtein distance of all label sequences to the predictions, normalised to the total length of all label sequences.
The predictions are obtained by using best path decoding~\cite{graves2006connectionist}, i.e. choosing the phoneme class with the highest probability at each time step.

To build the supervised and unsupervised training sets we keep all input data for unsupervised training and reduce the supervised set by drawing samples from the full dataset until the least represented phonemes are drawn a minimum number of times to prevent under-representing a class while keeping the distribution intact.
We cycle the supervised dataset to match sizes with the unsupervised set, similar to the implementation by Rasmus et al.~\cite{rasmus2015semi}.

\subsection{Training Procedure}

All networks have been trained using Adam~\cite{kingma2014adam} with a learning rate of $0.002$ for at least 100 epochs until the validation error stopped improving.
The models are four-layer networks consisting of one GRU layer with 192 units with $\text{tanh}(\cdot)$ activation and one feed-forward output layer with softmax activation, as well as the inverse layers in the decoder. The noisy softmax output is used to classify phonemes during training for additional regularisation.
Since the performance of the encoder is likely to correlate with RLN performance, hyper-parameters, including layer sizes and learning rate, were determined empirically by a grid search using the encoder described in section \ref{sec:rln}, i.e. an RLN with $\lambda_i = 0$, which also serves as the baseline. DAE cost weights $(\lambda_0,\lambda_1,\lambda_2) = (1000,10,0.1)$ and the MLP combinator $g(\cdot,\cdot)$ were both adopted from Rasmus et al.~\cite{rasmus2015semi}.

We test the semi-supervised learning capabilities of the six RLN variants from section \ref{sec:rln}, by varying the labels for the supervised part of the architecture in steps of 25\% (940), 50\% (1856), 75\% (2754), and 100\% (3696) of labelled sequences while the unsupervised part of the model always receives all available unlabelled data.

\section{Results \& Discussion}

\begin{figure}
    \centering
	\hspace{-1.em}
    \includegraphics[width=\textwidth]{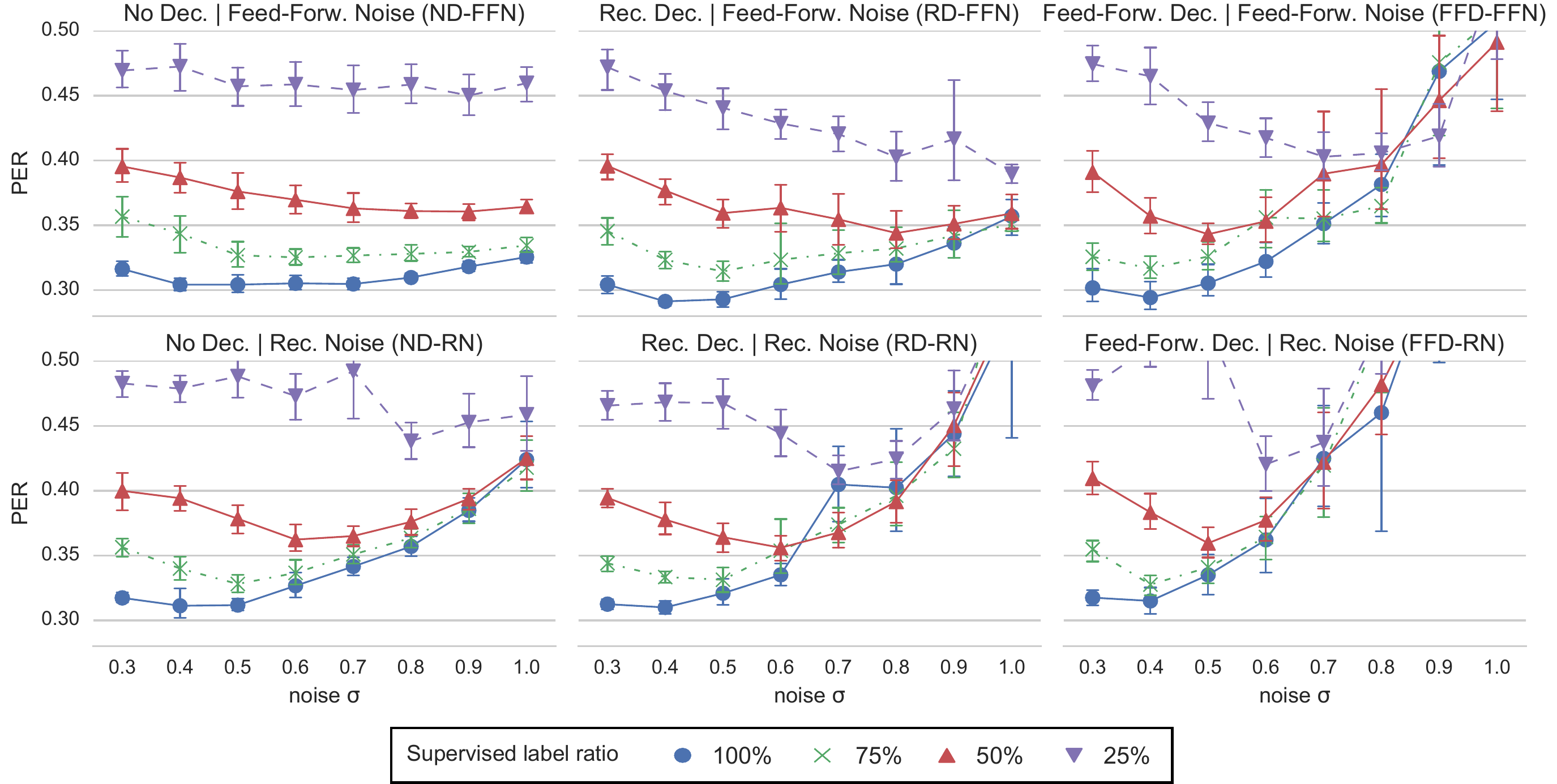}
    \caption{Comparison of PER achieved by the RLN variants with varying amount of labelled data (25\%-100\%) and noise standard deviation $\sigma$. Each data point represents the mean, the whiskers cover a 95\% confidence interval. Higher $\sigma$ are needed for fewer labels to prevent overfitting.}
    \label{fig:noises}
\end{figure}

An overview of our results can be seen in Fig. \ref{fig:noises} where the different modelling choices are directly compared against each other.
The overall best results after hyper-parameter optimization for each supervised data split, as well as results of other approaches, are shown in Table \ref{tab:results}.

As can be seen in Table \ref{tab:results}, the RLN consistently outperforms the baseline configuration, even in fully-supervised training and is able to achieve the same performance as the baseline with 25\% less labelled data which shows that the RLN complements the encoder well and demonstrates the compatibility of the LN with existing models.
On average, the RD models perform better than the FFD models for most $\sigma$, more so with fewer labels, suggesting that the recurrent decoder is better at filtering noise. This also explains why the RD models work better with higher $\sigma$ compared to FFD.

The noise injection method and the chosen $\sigma$ greatly impact the overall performance.
The performance curves are roughly concave and shift towards stronger noise with less available labels because the network overfits easily with fewer labels which is prevented by the higher noise.
Performance degrades for higher $\sigma$ because the network needs to be trained significantly longer to remove the noise which the chosen training parameters do not allow.

Recurrent noise injection was expected to achieve better regularisation due to the additional noise at the recurrent layer but does not.
By observing the encoder layers we found that their outputs often differed significantly which causes unrecoverable perturbations in the recurrent layers when applying equally strong noise to all layers instead of noise relative to each layer's output.
Employing batch normalisation might solve this, as hypothesised in related work~\cite{zhang2016augmenting}: normalising the preactivation of each layer to unit variance before adding noise makes the change in variance relative to the preactivation, therefore coupling noise and layer activation strength with the benefit of reducing the search space for $\sigma$ significantly. We predict that this will lead to an increase in performance when using fewer labels.

Even though our best results for the RLN are slightly lower ranked when compared with related approaches, our model has significantly fewer parameters (e.g. differing by a factor of 160 when compared to SSSAE~\cite{dhaka2016semi}). We therefore hypothesise that an increase of parameters and more complex layer architectures will result in even better performance.
This is indicated by our best RLN achieving similar results (31.66\% PER, 175k parameters) as the Bi-directional Long Short-Term Memory (BLSTM) (\cite{graves2006connectionist}; 31.25\% PER, 114k parameters) while using only half of the labels.




\begin{table}
\caption{Best results in phoneme error rate (PER), achieved by the proposed RLN modelling options: No decoder (ND, baseline), recurrent decoder (RD), feedforward decoder (FFD), feedforward noise (FFN), and recurrent noise (RN). $\dagger$: linear interpolation between 10\% and 30\% labels. $\dagger\dagger$: Graves et al.~\cite{graves2013speech} have shown significantly improved results with more parameters (17.7\% PER, 4.3\unit{m} param.).}%
\label{tab:results}%
\vspace{.5em}%
\centering%
\newcolumntype{C}{>{\centering\arraybackslash}X}%
\begin{tabularx}{\textwidth}{@{}r|ccc|ccc|cC@{}}
labels & ND-FFN & RD-FFN & FFD-FFN & ND-RN & RD-RN & FFD-RN & SSSAE \cite{dhaka2016semi} & BLSTM \cite{graves2006connectionist} \\
\hline%
25\% & 40.65 & \textbf{36.40} & 37.13 & 39.90 & 38.82 & \textbf{36.41} & 31.0$^\dagger$ & - \\
50\% & 34.22 & \textbf{31.66} & 32.06 & 34.07 & \textbf{33.07} & 33.39 & - & - \\
75\% & 30.96 & \textbf{29.16} & 30.31 & 31.17 &         30.84  & \textbf{30.42} & - & - \\
100\% & 29.11 & \textbf{28.02} & 28.08 & 29.26 &         29.67  & \textbf{29.26} & - & 31.25$^{\dagger\dagger}$ \\
\hline%
\hline%
param & 0.134\unit{m} & 0.177\unit{m} & 0.159\unit{m} & 0.134\unit{m} & 0.177\unit{m} & 0.159\unit{m} & 28.7\unit{m} & 0.114\unit{m} \\
\hline%
\end{tabularx}
\end{table}

\section{Conclusion}
%

%
We have shown that the recurrent ladder network is able to perform as good as similarly parametrised BLSTM models while using only 50\% of the labelled data, demonstrating the RLN's ability to effectively regularise itself using unsupervised training data.
Current state-of-the-art methods performed better overall but this does not come as a surprise given that these models use up to 160 times more parameters.
We argue that this gap could potentially be closed by scaling up our models, as demonstrated for BLSTM models by Graves et al.~\cite{graves2013speech}.

The proposed recurrent decoder proved to be better at denoising than the feed-forward decoder. Additionally, we found that recurrent noise injection does not perform as expected and we hypothesise that it needs the help of normalisation (e.g. batch normalisation) to work efficiently.

In the future, we would also like to take advantage of the semi-supervised learning abilities of the RLN in conjunction with more complex recurrent models such as bidirectional and attention-based RNNs to utilise unlabelled data even more effectively and explore how the learning framework scales with more complex temporal dynamics in more challenging tasks such as end-to-end speech recognition or question answering.





\subsubsection*{Acknowledgments.} The authors gratefully acknowledge partial support from the German Research Foundation DFG under project CML (TRR 169), the European Union under project SECURE (No 642667), and the Hamburg Landesforschungsf\"orderungsprojekt CROSS.

\bibliographystyle{splncs03} 


\end{document}